\documentclass[letterpaper, 10 pt, conference]{ieeeconf}  

\IEEEoverridecommandlockouts                              

\usepackage[utf8]{inputenc} 
\usepackage[T1]{fontenc}    
\usepackage{hyperref}       
\usepackage{url}            
\usepackage{booktabs}       
\usepackage{amsfonts}       
\usepackage{nicefrac}       
\usepackage{microtype}      
\usepackage{xcolor}         

\usepackage{bm}
\usepackage{amsmath}
\usepackage{amssymb}
\usepackage{mathtools}
\usepackage{amsthm}
\usepackage{caption}
\usepackage{subfigure}
\usepackage{multirow}
\usepackage{makecell}
\usepackage{fontawesome}
\usepackage{marvosym}
\usepackage[top=57pt, bottom=43pt, left=48pt,right=48pt]{geometry}
\usepackage{graphicx}
\usepackage{wrapfig}

\newcommand{\name}{{\textrm{DualCL}}}

\usepackage[ruled,vlined]{algorithm2e}
\usepackage{algorithmicx}
\usepackage{algpseudocode}
\usepackage{makecell}

\title{\LARGE \bf
A Dual Curriculum Learning Framework for Multi-UAV Pursuit-Evasion in Diverse Environments 
}

\author{Jiayu Chen$^{1}$,
Guosheng Li$^{1}$,
Chao Yu$^{1}$\textsuperscript{\Letter},
Xinyi Yang$^{1}$,
Botian Xu$^{*}$,
Huazhong Yang$^{1}$,
Yu Wang$^{1}$\textsuperscript{\Letter}
\thanks{{\Letter} Corresponding Authors. \url{{yuchao,yu-wang}@tsinghua.edu.cn}}
\thanks{$^{1}$Department of Electronic Engineering, Tsinghua University, Beijing, 100084, China.}
\thanks{$^{*}$ Work done as an intern in Tsinghua University}
\thanks{This research was supported by National Natural Science Foundation of China (No.62325405, U19B2019, M-0248), Tsinghua University Initiative Scientific Research Program, Tsinghua-Meituan Joint Institute for Digital Life, Beijing National Research Center for Information Science, Technology (BNRist) and Beijing Innovation Center for Future Chips.}
}

\begin{document}

\maketitle
\thispagestyle{empty}
\pagestyle{empty}

\begin{abstract}
This paper addresses multi-UAV pursuit-evasion, where a group of drones cooperates to capture a fast evader in a confined environment with obstacles. Existing heuristic algorithms, which simplify the pursuit-evasion problem, often lack expressive coordination strategies and struggle to capture the evader in extreme scenarios, such as when the evader moves at high speeds. In contrast, reinforcement learning (RL) has been applied to this problem and has the potential to obtain highly cooperative capture strategies. However, RL-based methods face challenges in training for complex 3-dimensional scenarios with diverse task settings due to the vast exploration space. The dynamics constraints of drones further restrict the ability of reinforcement learning to acquire high-performance capture strategies. In this work, we introduce a dual curriculum learning framework, named {\name},  which addresses multi-UAV pursuit-evasion in diverse environments and demonstrates zero-shot transfer ability to unseen scenarios. {\name} comprises two main components: the \emph{Intrinsic Parameter Curriculum Proposer}, which progressively suggests intrinsic parameters from easy to hard to continually improve the capture capability of drones, and the \emph{External Environment Generator}, tasked with exploring unresolved scenarios and generating appropriate training distributions of external environment parameters to further enhance the capture performance of the policy across various scenarios. The simulation experimental results show that {\name} significantly outperforms baseline methods, achieving over $90\%$ capture rate and reducing the capture timestep by at least $27.5\%$ in the training scenarios. Additionally, it exhibits the best zero-shot generalization ability in unseen environments. Moreover, we demonstrate the transferability of our pursuit strategy from simulation to real-world environments. Further details can be found on the project website at \url{https://sites.google.com/view/dualcl}.
\end{abstract}

\section{INTRODUCTION}\label{sec:intro}
The multi-UAV pursuit-evasion involves multiple drones cooperating to capture one fast evader within a confined environment with obstacles. In the domain of unmanned aerial vehicles (UAVs), pursuit-evasion tasks serve as critical platforms for exploring cooperative autonomy and motion planning strategies. The pursuit-evasion problem mirrors real-world challenges like missile interception~\cite{turetsky2003missile}, aircraft control~\cite{eklund2011switched}, criminal pursuit, as well as search and rescue missions~\cite{oyler2016pursuit}. The main problem is how multiple drones can cooperate to capture the evader efficiently in a complex environment with obstacles.

Heuristic algorithms have been widely used for pursuit-evasion problems~\cite{janosov2017group, angelani2012collective, fang2020cooperative}. Heuristic approaches require no training and can be directly applied to different scenarios. However, these methods often simplify the pursuit-evasion problem, such as using a fixed-speed mathematical model for pursuers, which cannot express complex pursuit strategies. Directly applying heuristic algorithms to UAVs, which are pursuers with dynamics constraints, often yields suboptimal results. Thus, the performance declines sharply as the task conditions become challenging, such as when the capture radius is very small or when the evader's velocity is very high. 

Reinforcement learning (RL), in contrast, can obtain pursuit strategies that are hard to encode through explicit rules, serving as a promising approach for pursuit-evasion problems~\cite{gupta2017cooperative, wang2019largescale, desouky2011q, awheda2015residual}. However, directly employing reinforcement learning to tackle the multi-UAV pursuit-evasion problem can be difficult. On the one hand, the 3-dimensional state space introduces a vast exploration space, further complicated by different environmental conditions such as different positions and quantities of obstacles, posing a significant challenge in training the policy using reinforcement learning. The vast exploration space leads to extensive training time for obtaining approximate optimal RL policies for diverse scenarios, potentially resulting in only suboptimal solutions. Most RL-based methods focus on the 2-dimensional cases and only consider fixed task settings~\cite{gupta2017cooperative, wang2020cooperative, zhang2022multi}. On the other hand, we need to model the dynamics of drones to ensure the effective transfer of RL policies from the simulation to real-world scenarios. Drones, unlike particle agents, are subjected to dynamics constraints, further limiting their behaviors and making it challenging to explore optimal strategies.

To address multi-UAV pursuit-evasion in diverse environments, we propose a dual curriculum learning method ({\name}), which utilizes curriculum learning to accelerate training and explore effective pursuit policy. Our method consists of two main components, an \emph{Intrinsic Parameter Curriculum Proposer} and an \emph{External Environment Generator}. The \emph{Intrinsic Parameter Curriculum Proposer} adjusts intrinsic parameters to propose new tasks gradually increasing in difficulty, guiding the progression of the pursuit policy. The \emph{External Environment Generator} efficiently explores and preserves unsolved scenarios, focusing on learning from these scenarios through resampling to enhance the ability to handle complex cases. We compare {\name} with existing works in the NVIDIA Isaac Sim-based Omnidrones environment (Omni)~\cite{xu2023omnidrones}. We design six test scenarios to evaluate the ability of algorithms to handle hard scenarios and the zero-shot transfer ability. In \emph{Empty}, our method achieves almost $100\%$ capture rate while all baselines completely fail. In \emph{1-Central Tower}, \emph{2-Curve Search} and \emph{3-Central Tower}, {\name} achieves over $90\%$ capture rate and reduces the capture timestep by at least $27.5\%$. In \emph{4-Curve Search} and \emph{5-Central Tower}, {\name} outperforms all baseline algorithms, showing the best out-of-distribution generalization. We also provide a real-world demonstration to illustrate the feasibility of executing the capture policy generated by {\name}.

\begin{figure*}[t]
    \centering
    \includegraphics[width=0.92\textwidth]{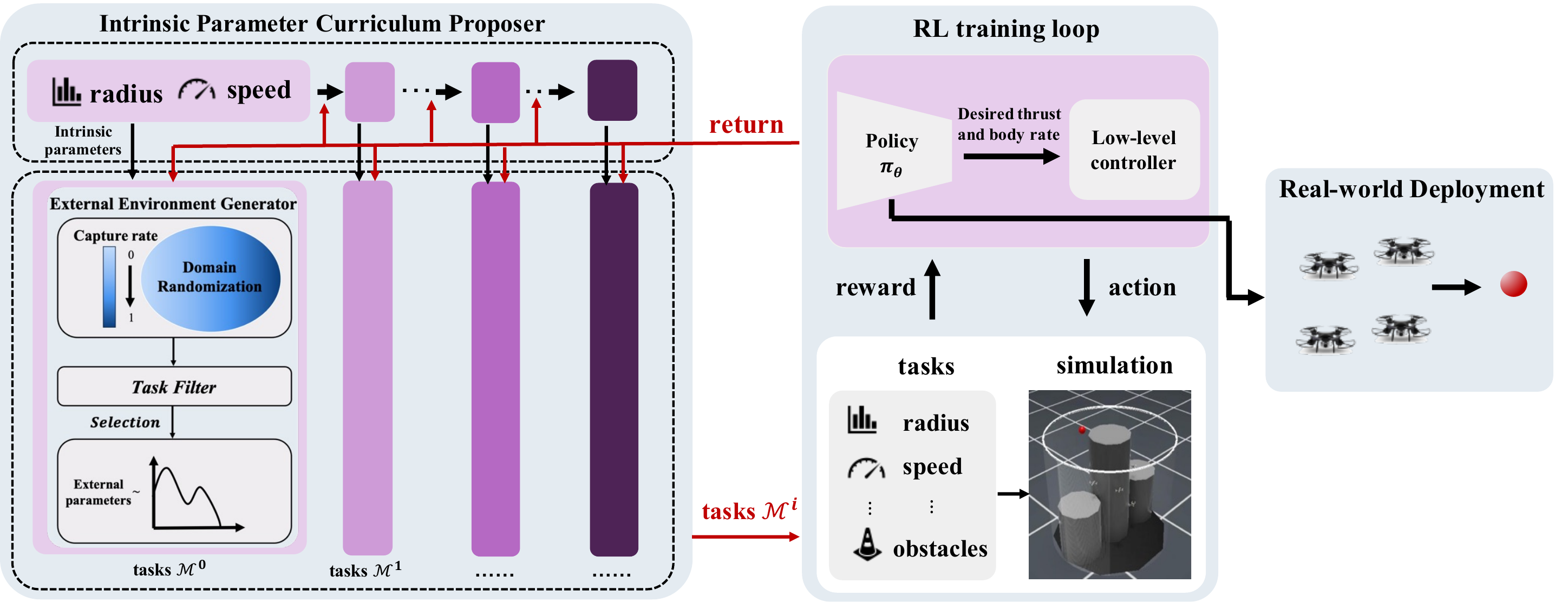}
    \caption{Workflow of our dual curriculum learning framework. Taking the phase $i$ as an example, we utilize the \emph{Intrinsic Parameter Curriculum Proposer} to generate the intrinsic parameters $\tau^i_{int}$ for the current phase and the \emph{External Environment Generator} to produce external environment parameters $\tau^i_{ext}$. We combine the intrinsic and external environment parameters to form the task parameters set $\mathcal{M}^i$ and generate parallel simulation environments for MARL training.}
    \vspace{-3mm}
    \label{fig:overview}
\end{figure*}

\section{RELATED WORK}\label{sec:related}
\subsection{Multi-UAV Pursuit-Evasion}
There are two kinds of methods for pursuit-evasion problems: heuristic pursuit algorithms and reinforcement learning-based algorithms. Heuristic algorithms design the mathematical forms of force to guide the pursuers' move~\cite{janosov2017group, angelani2012collective, fang2020cooperative,muro2011wolf}. For example, ~\cite{janosov2017group} assumes that each pursuer is attracted by the evader and moves toward the evader while being repelled by its teammates and obstacles to prevent collisions. ~\cite{fang2020cooperative} introduces a surrounding force that enables pursuers to form an encirclement and approach the faster evader. We can employ heuristic algorithms as high-level planners, providing desired velocities as input to the lower-level controller to control drones, thereby straightforwardly extending these methods to multi-UAV pursuit-evasion problems. However, the pursuers in these methods are defined by fixed-speed mathematical models that chase the evader only by controlling the motion direction without considering the cooperativeness among multiple pursuers. 

Another promising method for pursuit-evasion is via deep RL, which learns cooperative pursuit strategies~\cite{wang2020cooperative, gupta2017cooperative,xu2020multi, huttenrauch2019deep}. For example, ~\cite{gupta2017cooperative} learns a pursuit strategy using a parameter-sharing Trust Region Policy Optimization (TRPO) algorithm that coordinates multiple pursuers to capture the evader. ~\cite{wang2020cooperative} applies Multi-agent Deep Deterministic Policy Gradient (MADDPG)~\cite{lowe2017multi} to learn cooperative strategies to chase an evader. ~\cite{xu2020multi} focuses on pursuit-evader games with non-holonomic agents, where new agents can join the game. ~\cite{huttenrauch2019deep} proposes a new state representation for deep multi-agent RL training by considering the agents as interchangeable and the exact number irrelevant. However, these methods consider pursuit-evasion problems in 2-dimensional spaces and have not been extended to cover 3-dimensional multi-UAV scenarios due to the exponentially expanding state space. Considering the dynamics constraints of multi quadcopter, \cite{zhang2022game} design a target prediction network (TP Net) combined with MADDG to generate pursuit strategies for multi-UAV pursuit-evasion. However, they constrain the state space of drones to a 2-dimensional plane, essentially treating this problem as a 2-dimensional pursuit-evasion problem. Moreover, these works primarily focus on solving pursuit-evasion tasks within a fixed environment. Our approach, on the other hand, investigates how to concurrently solve three-dimensional multi-UAV pursuit-evasion tasks in diverse scenarios, taking into account the dynamics constraints of UAVs.

\subsection{Curriculum Learning}
Curriculum learning (CL)~\cite{bengio2009curriculum} is a learning paradigm that accelerates training by gradually increasing the difficulty of the training tasks. Automatic curriculum learning (ACL)~\cite{portelas2020automatic, florensa2017reverse, florensa2018automatic, ivanovic2019barc} is a method that evaluates the difficulty of a task by some metrics, such as task success rate, and automatically generates the training distribution based on the current policy. The core idea of ACL is to train the policy using tasks with moderate difficulty, i.e., neither too hard nor too easy. In single-agent tasks, ~\cite{florensa2017reverse, florensa2018automatic} assume a fixed goal and generate tasks with starting states increasingly far away from the goal to obtain a curriculum. ~\cite{akkaya2019solving, mehta2020active} consider a large task space over simulator configurations and train from the most accessible settings to the hardest for sim-to-real adaptation. Recently, several works have applied ACL in multi-agent systems to efficiently train a cooperative policy~\cite{chen2021variational,long2020evolutionary,wang2019largescale}. ~\cite{chen2021variational} proposes a curriculum learning algorithm to tackle goal-conditioned cooperative MARL problems with sparse rewards through a novel variational inference perspective. \cite{long2020evolutionary,wang2019largescale} propose a population curriculum that trains entity-invariant policies over training tasks with a growing number of agents. Our algorithm proposes a dual curriculum learning framework that employs an \emph{Intrinsic Parameter Curriculum Proposer} to address the challenge of three-dimensional state space and an \emph{External Environment Generator} to explore nearly optimal policy across diverse environments.

\section{PRELIMINARY}\label{sec:prelim}
\subsection{Problem Definition}
In this paper, the multi-UAV pursuit-evasion problem is defined as N drones pursuing a faster evader in a scenario with obstacles. The goal of drones is to catch the evader as quickly as possible while avoiding obstacles. Conversely, the goal of the evader is to get as far away from the drones and obstacles as possible. The evader will be captured by a drone if the distance between them is less than $d_{c}$. Each episode of the capture contains $T$ timesteps. If the time $T$ runs out and the evader is still not captured, we define the capture in this episode as a failure. The following assumptions are further made:

The drones, the evader, and the obstacles are spawned randomly within an arena of a radius $R_a$. Drones and the evader can fly in the air up to a maximum height of $h$. Neither the drones nor the evader can get out of the arena. The maximum number of obstacles is denoted by 
$N_o$. Obstacles are modeled as cylinders with a radius of $r_c$ and a height ranging from $0$ to $h$. The collision radius of drones is $d_{col}$. The distance between drones and obstacles needs to be larger than $d_{col}$ to avoid collision. The maximum speed of the drone is $v_p$, and the evader maintains a constant speed $v_e$, where $v_p < v_e$.

We consider the evader as a part of the environment and use a potential field method~\cite{de2021decentralized} to control the evader. Drones and obstacles exert repulsion on the evader along the vector direction. The boundary also exerts a force perpendicular to the boundary on the evader so that the evader can avoid boundaries. These forces decrease with distance. The speed vector of the evader $\boldsymbol{v_e}$ is calculated by: 
\begin{equation}
\scalebox{0.9}{$
    \boldsymbol{v_e} = v_e \times \left(\sum_{i}
    \left(\frac{\boldsymbol{x}_{p,i} - \boldsymbol{x}_e}{{||\boldsymbol{x}_{p,i} - \boldsymbol{x}_e||}^2}\right)
    + \sum_{j}\left(\frac{\boldsymbol{d}_{sur,j}}{{||\boldsymbol{d}_{sur,j}||}^2}\right) + \sum_{k}\left(\frac{\boldsymbol{u}_k}{d_k}\right)\right)$
}
\end{equation}
where $\boldsymbol{x}_e$ is the current position of the evader, $\boldsymbol{x}_{p, i}$ is the current position of the drone $i$, $\boldsymbol{d}_{sur,j}$ is the distance vector from the closest point on the obstacle $j$ to the evader, pointing towards the evader. 
$\boldsymbol{u_k}$ represents the unit vector pointing from the point on the boundary nearest to the evader towards the evader. $d_k$ is the vertical distance from the evader to the boundary $k$. There are three boundaries, including the ground, the upper height limit, and the boundary of the arena.

\subsection{Multi-Agent Reinforcement Learning}

\textbf{Markov Game}
We formulate the multi-UAV pursuit-evasion problem as a multi-agent Markov decision process (MDP), $M=<\mathcal{N}, \Theta,\mathcal{S},\mathcal{A},\mathcal{O}, O, P, R,\gamma>$, where $\Theta$ represents the parameter set used to define the pursuit-evasion task. In our setup, $\Theta$ comprises the capture radius, the speed of the evader, the initial positions of the drones, and the evader, as well as the number, height, and positions of obstacles. $\mathcal{N}\equiv\{1,...,N\}$ is a set of $N=|\mathcal{N}|$ drones.  $\mathcal{S}$ is the state space. 
$\mathcal{A}$ is the shared action space for each drone. $\mathcal{O}$ is the observation space. $o_i=O(s;i)$ denotes the observation for the drone $i$ under state $s\in\mathcal{S}$. $P(s'|s,A)$ and $R(s,A)$ are the transition probability and the reward function given state $s$ and joint actions $A=(a_1,\ldots,a_N)$. $\gamma$ is the discount factor. 

We consider homogeneous drones and learn a shared policy $\pi_\theta(a_i|o_i)$ parameterized by $\theta$ for each drone $i$. The final objective $J(\theta)$ of the reinforcement learning algorithm is to maximize the expected accumulative reward for any task parameter $\tau\sim\Theta$, i.e.,
\begin{align}
J(\theta)=\mathbb{E}_{\tau\sim\Theta, a_i^t,s^t}\left[\sum_t \gamma^t R(s^t,A^t; \tau) \mid \tau \right].
\end{align}

\section{METHODOLOGY}\label{sec:method} 
\begin{algorithm}[t]
 \caption{Dual Curriculum Learning Framework}
 \label{algo:full}
   \textbf{Input:} {$\theta$, $B$, $\mathcal{Q}_{active}$, $d^{target}_c$, $v^{target}_e$, $N^{max}_o$, $h^{max}$}\; \tcp{$B:$ the batch size, $\mathcal{Q}_{active}:$ the active archive. Target superscripts denote the parameters of the target task setting.}
   
  \textbf{Output:} final policy $\pi_{\theta}$\;
  $i\gets 0, \mathcal{Q}_{valid}\gets \textbf{TaskFilter}(N^{max}_o, h^{max}),$ $\mathcal{Q}_{active}\gets\{\},  \mathcal{L}\gets\textbf{GetOrder}(d^{target}_c, v^{target}_e)$\;
  \Repeat{$i == |\mathcal{L}| - 1$}
  {
    \tcp{setup intrinsic parameters.}
    $\tau^{i}_{int} = [d^i_c, v^i_e]\gets \mathcal{L}[i]$\;
    \tcp{generate external parameters.}
    Sample $\tau^{i}_{ext} \sim \mathrm{Unif}(\mathcal{Q}_{active})$ with probability $p$, else $\tau^{i}_{ext} \sim \mathrm{Unif}(\mathcal{Q}_{valid})$ with probability $1- p$\;
    \tcp{Policy update.}
    $\mathcal{M}^i = \{\tau_j = (\tau^i_{int}, \tau^i_{ext}) | j = 0,1,…, B-1\}$\;
    Train and evaluate $\pi_{\theta}$ on $\mathcal{M}^i$ via MARL\;
    \tcp{Add unsolved tasks in $\mathcal{M}^i$ to the active archive.}
    $\mathcal{Q}_{active}\gets\textbf{Update}(\mathcal{Q}_{active}, \mathcal{M}^i)$\;
    
    \If{\textbf{Evaluation}($\tau^i_{int}, \pi_{\theta}$)}
    {$i = min(i + 1, |\mathcal{L}| - 1)$\;
    $\mathcal{Q}_{active}\gets\{\}$\;
    }
  }
 \end{algorithm}

\begin{figure*}[t]
\begin{minipage}{1.0\textwidth}
\subfigure[Infeasible scenarios.\label{fig:infeasible}]{\includegraphics[width=0.32\textwidth]{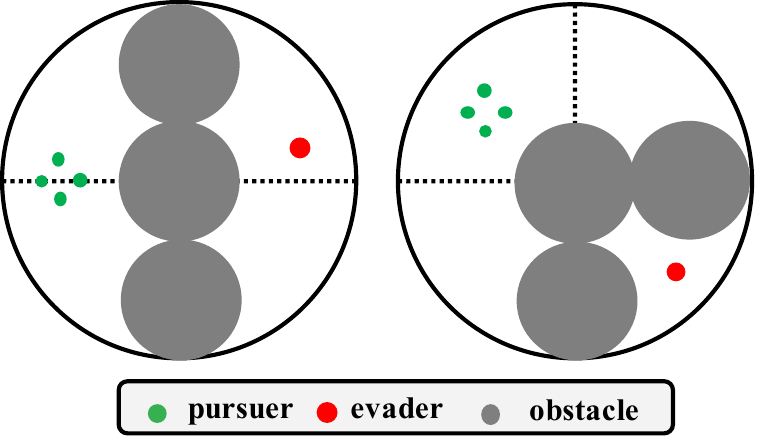}}
\subfigure[Feasible scenarios.\label{fig:feasible}]{\includegraphics[width=0.32\textwidth]{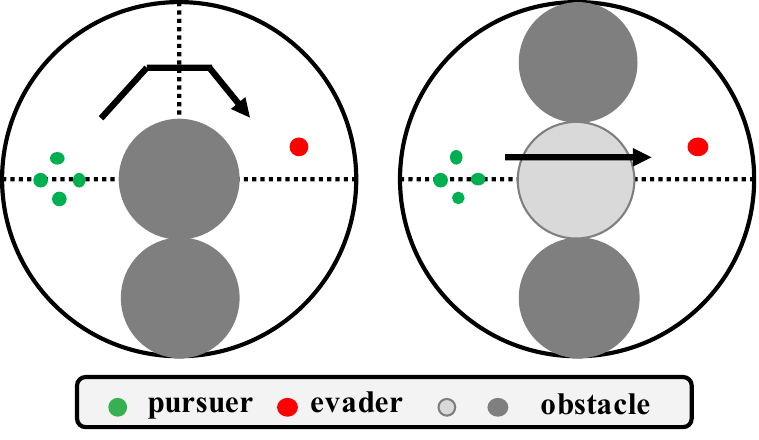}}
\subfigure[Depth-First Search.\label{fig:dfs}]
{\includegraphics[width=0.32\textwidth]{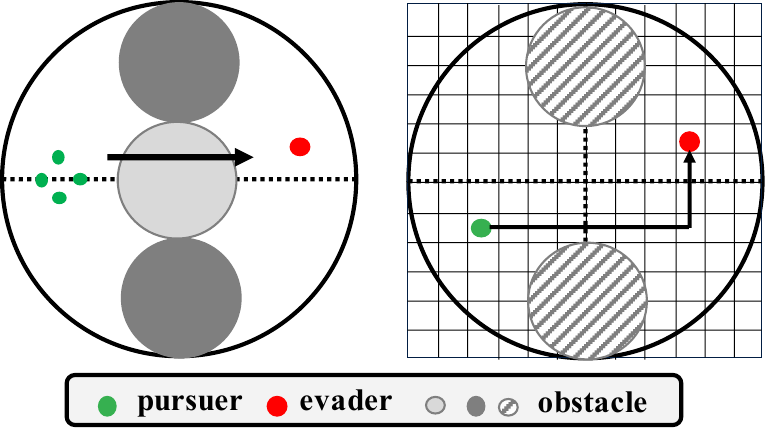}}
\caption{The aerial view of the multi-UAV pursuit-evasion problem. In (a) and (b), 
we demonstrate two infeasible scenarios and two feasible scenarios. In (c), we show the grid pattern of the scenario and verify its solvability using DFS.}\label{fig:scenarios}
\vspace{-3mm}
\end{minipage}
\end{figure*}

In this section, we first introduce the basic setup of MARL, which includes the observation space, action space, and reward function. Then, we introduce our dual curriculum learning framework, which consists of two main modules: the \emph{Intrisic Parameter Curriculum Proposer} and \emph{External Environment Generator}.

\subsection{Multi-Agent Reinforcement Learning Setup}

\subsubsection{Observation and Action Space}
The observations of drone $i$ include the self-state, the relative positions of other drones, the relative speed and position of the evader, the current time step, and the information on the obstacles. The self-state comprises velocity, position, and orientation represented with quaternion of the drone $i$. The information on the obstacles consists of the positions relative to the drone $i$, heights, and radius of all obstacles. We adopt a hierarchical control structure to drive the drones. At the top level, we employ an RL policy, which outputs $[c, \omega_d]$, comprising the collective thrust $c$ and the desired body angular rate $\omega_d$. We utilize a Proportional–integral–derivative (PID) controller as the lower-level controller, which computes thrust commands $u$ based on the collective thrust $c$ and the error between the actual angular rate $\omega$ and the desired angular rate $\omega_d$.

\subsubsection{Reward Design}

We consider a capture reward, $r_{cap}$, for successfully capturing the evader. If the distance between any drone and the evader falls below the catch radius, $d_c$, all drones are rewarded with a bonus of $10$. This reward encourages the pursuers to work collectively to capture the evader.

The collision penalty, $r_{col}$, acts as a critical safety measure to discourage drones from colliding with obstacles. When the distance between a drone $i$ and an obstacle is less than a threshold, $d_{col}$, the drone $i$ will receive a penalty of $-1$.

The overall reward, $r_{total}$, is the combination of the above two reward components to motivate drones to capture the evader while avoiding potential obstacles effectively. $r_{total}$ is expressed as:
\begin{align}
r_{total} = r_{cap} + r_{col}.
\end{align}

\subsection{Dual Curriculum Learning Framework}

As illustrated in Fig.~\ref{fig:overview}, we present the workflow of our dual curriculum learning framework designed for multi-UAV pursuit-evasion. This framework comprises two key components: an \emph{Intrinsic Parameter Curriculum Proposer}, which suggests intrinsic parameters to increase the pursuit's difficulty as training advances progressively, and an \emph{External Environment Generator}, which is employed to explore unsolved scenarios and proposing suitable external environmental parameters for policy improvement.

Initially, we categorize the task parameters $\tau$ into two groups, intrinsic and external environment parameters, and set the final target values for all task parameters. In our algorithm, the intrinsic parameters, $\tau_{int}$, consist of the capture radius $d_c$ and the speed of the evader $v_e$. The external environment parameters, $\tau_{ext}$, encompass the positions of drones and the evader, as well as the number, positions, and heights of obstacles. In the \emph{Intrinsic Parameter Curriculum Proposer}, we design a sequence for intrinsic parameters from easy to hard and propose intrinsic parameters from the sequence by the training progress. The \emph{External Environment Generator} filters out infeasible scenarios employing the \emph{TaskFilter} and establishes a valid environment parameter distribution $\mathcal{Q}_{valid}$. We select and preserve unsolved scenarios from $\mathcal{Q}_{valid}$ to form the active archive $\mathcal{Q}_{active}$ based on the episodic capture return. Next, the \emph{External Environment Generator} samples from both $\mathcal{Q}_{valid}$ and $\mathcal{Q}_{active}$ to obtain the external environment parameters $\tau_{ext}$. Finally, we combine the intrinsic and external environment parameters to obtain the task parameter set $\mathcal{M}$ and generate parallel simulation environments for MARL training. Note that this framework is compatible with any MARL algorithm. In this paper, we use a state-of-the-art MARL algorithm MAPPO~\cite{yu2022mappo} as our training backbone. A detailed pseudo code of this entire process can be found in Algo.~\ref{algo:full}.

\subsubsection{Intrinsic Parameter Curriculum Proposer}\label{sec:algo:outer}
The \emph{Intrinsic Parameter Curriculum Proposer} is designed to generate a sequence $\mathcal{L}$ of intrinsic parameters $\tau_{int}$ with gradually increasing difficulty and propose appropriate intrinsic parameters based on the training progress. We leverage an important conductive bias in the multi-UAV pursuit-evasion problem that the difficulty of capture increases as the capture radius decreases and the speed of the evader increases due to space constraints. The most challenging scenario arises when the drones are required to approach an extremely fast evader infinitely closely. We initially set the catch radius to the radius of the arena and the initial speed of the evader to $0.0$, which defines the simplest pursuit-evasion task. Given
the target values of the intrinsic parameters, we define the sequence $\mathcal{L}$ as follows: divide the interval between the initial and target values into ten equal parts to obtain sequences for the two intrinsic parameters. We start by setting the catch radius to the initial value and iterate over the sequence of the speed of the evader. Once the speed of the evader reaches the target value, we begin iterating over the sequence of the catch radius until both intrinsic parameters reach the target values. For phase $i\in[0, |\mathcal{L} - 1|]$, we set the intrinsic parameters to $i$th value in $\mathcal{L}$. For policy advancement, we evaluate the policy $\pi_{\theta}$ in the simulation with the intrinsic parameter $\tau^i_{int}$ and compute the average capture rate of 1000 episodes using the episodic capture return $\sum_{t}r_{cap,t}$. If the average capture rate is equal to or greater than the success threshold $\sigma$, we can advance the policy from phase $i$ to phase $i+1$, i.e.,
\begin{equation}
Evaluation(\tau^i_{int}, \pi_{\theta}) = mean(\sum_{t} r_{cap, t} > 0) \geq \sigma
\end{equation}
In our experiments, we set $\sigma = 0.98$.

\subsubsection{External Environment Generator}\label{sec:algo:generator}
In this section, we introduce the \emph{External Environment Generator}, a pivotal component that efficiently obtains the suitable training distribution of external environment parameters. The \emph{External Environment Generator} consists of two parts: the \emph{Task Filter} and \emph{Selection}. 

The \emph{Task Filter} is responsible for generating feasible scenarios. To prevent the policy from overfitting to a single scenario, we adopt Domain Randomization (DR)~\cite{tobin2017domain} to generate drones and the evader at different positions, as well as obstacles of varying heights, positions, and quantities. However, DR can lead to infeasible scenarios, resulting in degraded capture strategies. For example, as shown in Fig.~\ref{fig:infeasible} and Fig.~\ref{fig:feasible}, we demonstrate infeasible scenarios and feasible scenarios where dark gray obstacles are at the same height as the arena, preventing passage over them. Light gray obstacles are lower than the arena, allowing pursuers and the evader to pass over them from above. As shown in Fig.~\ref{fig:dfs}, we employ Depth-First Search (DFS) to filter out infeasible scenarios. We grid the scenarios into 2-dimensional grids, with each grid being the same size as the drones. For randomly generated obstacles, we mark the corresponding positions within the grid as occupied if the positions of obstacles are not passable in 3-dimensional space. Thus, in Fig.~\ref{fig:dfs}, the dark gray obstacles at the top and bottom are impassable, while the light gray obstacle in the middle is unoccupied in the grid. Taking the drones' initial position as the starting point and the evader's initial position as the endpoint, we use DFS to determine whether a feasible path exists from the starting point to the endpoint. We define a scenario as solvable if a feasible path exists between all drones and the evader. We define the filtered external environment parameter distribution as $\mathcal{Q}_{valid}$.

The \emph{Selection} is responsible for exploring unsolved scenarios under the distribution $\mathcal{Q}_{valid}$. We maintain an active archive $\mathcal{Q}_{active}$ to store scenarios that the current policy still needs to solve. We define the updating process of $\mathcal{Q}_{active}$ as \emph{Update}. Specifically, tasking the phase $i$ as an example, we generate parallel simulation environments for MARL training according to the task parameter set $\mathcal{M}_i$ and obtain the episodic capture return corresponding to the external environment parameter $\tau^i_{ext}$. We include the external environment parameters satisfying the criterion into the active archive $\mathcal{Q}_{active}$, i.e.,

\begin{equation}
        \mathcal{Q}_{active}\gets\{\tau^i_{ext, j} | \sum_{t}r_{cap, t} = 0, j\in [0, |\mathcal{M}_i| - 1] \},
\end{equation}

We use a fixed-size queue to maintain $\mathcal{Q}_{active}$. When the size of $\mathcal{Q}_{active}$ exceeds the capacity, we randomly remove the old elements for update. The updating process of $\mathcal{Q}_{active}$ continuously increases the complexity of tasks based on the current policy, leading to automatic curriculum generation.

Finally, we select external environment parameters from the active archive $\mathcal{Q}_{active}$ with probability $p$, while those from $\mathcal{Q}_{valid}$ are selected with probability $1-p$. In our experiments, we choose $p = 0.7$. Our choice to sample from $\mathcal{Q}_{active}$ is driven by the objective of exploring unsolved scenarios and accelerating convergence through the utilization of tasks of moderate complexity. Conversely, sampling from the valid distribution, $\mathcal{Q}_{valid}$, is essential for promoting global exploration and preventing the agents from forgetting previously acquired knowledge and skills.

\vspace{3mm}
\section{EXPERIMENT}\label{sec:expr}
\begin{figure*}[t]
\centering
\begin{minipage}{\textwidth}
\begin{subfigure}
    \centering
    \includegraphics[width=\textwidth]{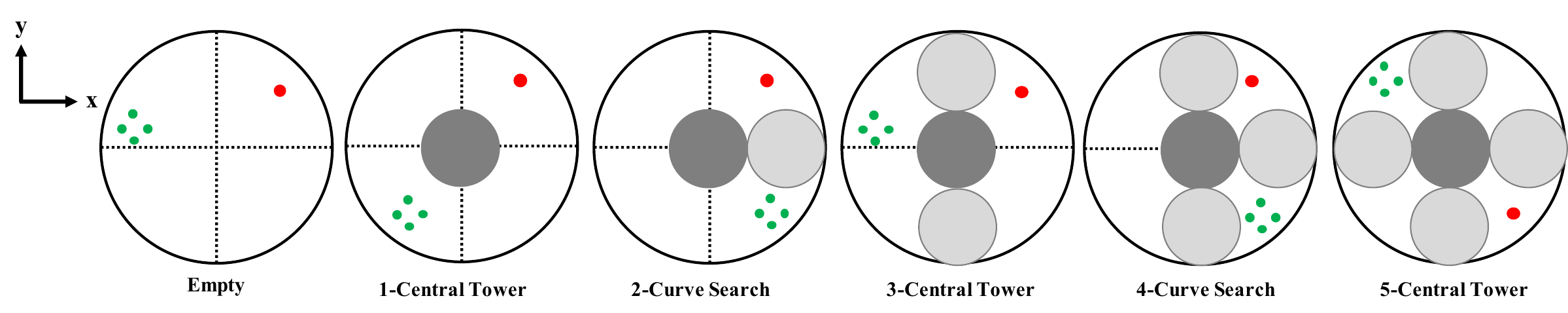}
    \caption{The test scenarios. The \emph{Empty}, \emph{1-Central Tower}, \emph{2-Curve Search} and \emph{3-Central Tower} are designed to assess the algorithms' cooperativeness and the ability to tackle challenging environments. The \emph{4-Curve Search} and \emph{5-Central Tower} are specifically chosen to evaluate the zero-shot transfer capability of algorithms.}
    \label{fig:evaluation}
\end{subfigure}
\end{minipage}
\end{figure*}

\subsection{Task Setting}
In the training setup, we tune the hyperparameters of heuristic algorithms and train RL-based algorithms within an arena with a radius $R_a$ of $0.9$ and a maximum height $h$ constraint of $1.2$. The maximum speed of the drones $v_p$ is set to $1.0$. The size of drones is set to $0.1$. The target capture radius $d^{target}_c$ is set to $0.12$, which is slightly larger than the size of a drone, with the evader's target speed $v^{target}_e$ established at $2.4$. Each obstacle has a radius $r_c$ of $0.3$ and a maximum height of 1.2, with their quantity ranging from $0$ to $3$. Both drones and the evader are randomly generated within the arena. The cylinders are randomly generated within the arena, while the height of each cylinder is randomly chosen from either $0.6$ or $1.2$. Each episode consists of $800$ timesteps.

\subsection{Test Scenarios}
As illustrated in Fig.~\ref{fig:evaluation}, we present aerial views of the six test scenarios.  The \emph{Empty}, \emph{1-Central Tower}, \emph{2-Curve Search} and \emph{3-Central Tower} which are within-distribution scenarios serve to show the algorithm's cooperativeness and the ability to handle challenging scenarios. We also design two out-of-distribution scenarios, \emph{4-Curve Search} and \emph{5-Central Tower} to evaluate the zero-shot transfer capability of algorithms. The obstacles in dark gray have a height of $1.2$, matching the arena's height, making them impassable from above. On the other hand, the light gray obstacles have a height of $0.6$, allowing both drones and the evader to pass over them from above.

\subsection{Evaluation Metrics}
We consider two statistical metrics to capture different characteristics of a particular pursuit strategy. Every experiment is repeated over three seeds. Each evaluation score is expressed in the format of "mean (standard deviation)", which is averaged over a total of 3000 testing episodes, i.e., 1000 episodes per random seed.
\begin{itemize}
    \item \textbf{Capture rate: }This metric shows the average success rate of capturing the evader in an episode. We define a successful capture as the evader being captured before the time $T$ runs out. Higher \emph{capture rate} indicates stronger capture abilities. 
    \item \textbf{Capture timestep: }This metric represents the average timestep when the evader is captured in an episode. If a capture is failed, the capture timestep is defined as the maximum length of an episode. Lower \emph{capture timestep} indicates faster capture.
\end{itemize}

\begin{table*}[t]
\centering
\footnotesize
{
{\begin{tabular}{ccccc|c|cc} 
\toprule
Scenarios& Metrics&Angelani&Janosov &APF  & DACOOP &MAPPO + \emph{Intrinsic} & {\name}  \\ 
\midrule
\multirow{3}{*}{\emph{1-Central Tower}} & \textit{Cap. Rate}$\uparrow$ & 0.043\scriptsize{(0.019)} & 0.013\scriptsize{(0.005)} & 0.013\scriptsize{(0.005)} & 0.063\scriptsize{(0.005)} & 0.703\scriptsize{(0.082)} &   \textbf{0.934\scriptsize{(0.019)}}     \\ 
\cmidrule{2-8}
                     & \textit{Cap. Time}$\downarrow$  & 774.3\scriptsize{(006.6)} & 790.0\scriptsize{(002.8)} & 790.0\scriptsize{(002.8)}   & 760.7\scriptsize{(003.1)}& 561.5\scriptsize{(32.37)} &    \textbf{407.0\scriptsize{(015.3)}}    \\ 
\cmidrule{1-8}
\multirow{3}{*}{\emph{2-Curve Search}} & \textit{Cap. Rate}$\uparrow$ & 0.370\scriptsize{(0.045)} &0.393\scriptsize{(0.034)} & 0.393\scriptsize{(0.034)} &0.513\scriptsize{(0.043)} & 0.167\scriptsize{(0.030)} &   \textbf{0.935\scriptsize{(0.007)}}     \\ 
\cmidrule{2-8}
                     & \textit{Cap. Time}$\downarrow$   & 672.7\scriptsize{(024.2)} & 650.3\scriptsize{(020.1)} &  650.3\scriptsize{(020.1)}   & 552.3\scriptsize{(018.6)}& 744.7\scriptsize{(007.0)} &   \textbf{398.1\scriptsize{(008.9)}}    \\ 
\cmidrule{1-8}
\multirow{3}{*}{\emph{3-Central Tower}} & \textit{Cap. Rate}$\uparrow$ & 0.533\scriptsize{(0.037)}  & 0.559\scriptsize{(0.025)}& 0.460\scriptsize{(0.041)}  &0.631\scriptsize{(0.009)} & 0.725\scriptsize{(0.029)}  &   \textbf{0.961\scriptsize{(0.024)}}     \\ 
\cmidrule{2-8}
                     & \textit{Cap. Time}$\downarrow$   & 571.7\scriptsize{(017.8)}   & 559.3\scriptsize{(014.8)}   & 583.7\scriptsize{(008.6)} &525.3.0\scriptsize{(007.7)} & 513.6\scriptsize{(15.87)} & \textbf{362.0\scriptsize{(41.40)}}    \\ 
\cmidrule{1-8}
\multirow{3}{*}{\emph{4-Curve Search}} & \textit{Cap. Rate}$\uparrow$ & 0.507\scriptsize{(0.054)} & 0.767\scriptsize{(0.041)} & 0.560\scriptsize{(0.043)} &0.782\scriptsize{(0.034)} & 0.563\scriptsize{(0.036)} &   \textbf{0.931\scriptsize{(0.015)}}     \\ 
\cmidrule{2-8}
                     & \textit{Cap. Time}$\downarrow$   &  610.7\scriptsize{(021.9)} & 442.3\scriptsize{(017.3)} & 554.7\scriptsize{(024.5)} & 414.7\scriptsize{(016.5)} & 598.4\scriptsize{(09.67)}&\textbf{383.5\scriptsize{(009.7)}}    \\ 
\cmidrule{1-8}
\multirow{3}{*}{\emph{5-Central Tower}} & \textit{Cap. Rate}$\uparrow$ & 0.437\scriptsize{(0.063)} &0.723\scriptsize{(0.026)} &0.773\scriptsize{(0.024)}  &0.843\scriptsize{(0.024)} & 0.519\scriptsize{(0.128)} &   \textbf{0.886\scriptsize{(0.030)}}     \\ 
\cmidrule{2-8}
                     & \textit{Cap. Time}$\downarrow$   & 632.0\scriptsize{(032.8)} & 499.3\scriptsize{(013.1)}   & 512.0\scriptsize{(009.9)} & {492.7\scriptsize{(006.9)}} & 617.1\scriptsize{(69.15)} &\textbf{476.7\scriptsize{(026.1)}}    \\ 
\bottomrule
\end{tabular}}}
\caption{Results of {\name} and baselines in the scenarios with obstacles. {\name} outperforms both heuristic and RL-based methods and shows the best out-of-distribution generalization.}
\label{tab:main}
\end{table*}


\subsection{Baselines}
We challenge {\name} with three heuristic algorithms (Angelani, Janosov, and APF) and two RL-based algorithms (DACOOP and MAPPO).

\begin{itemize}
    \item \textbf{Angelani}~\cite{angelani2012collective}: Angelani describes a pursuit-evasion phenomenon arising from intergroup interactions, where pursuers are attracted to the nearest particles of the opposing group, i.e., the evader.
    \item \textbf{Janosov}~\cite{janosov2017group}: Following the concepts of the Vicsek particle model~\cite{angelani2012collective}, Janosov designs greedy chasing of evaders and collision avoidance in a realistic scenario with inertia, time delay and noise.
    \item \textbf{APF}~\cite{koren1991potential}: APF guides pursuers to a target position via the combination of attractive, repulsive, and inter-individual forces. By setting the target of pursuers to the evader's position and adjusting the hyperparameters for different forces, pursuers can move towards the evader with obstacle avoidance in APF. 
    \item
    \textbf{DACOOP}~\cite{zhang2022multi}: DACOOP is an RL-based adaptive cooperative pursuit algorithm that combines RL and APF. DACOOP uses RL to modify the primary hyperparameters of APF and adapt them to different task conditions.
    \item
    \textbf{MAPPO}~\cite{yu2022mappo}:  MAPPO is one of the most advanced multi-agent reinforcement learning algorithms for general cooperative problems. We consider MAPPO to be a naive baseline for policy training without any curriculum learning techniques.
\end{itemize}

For the heuristic algorithms, we grid-search the hyperparameters of the algorithm in the training scenarios and select the best results. For the RL-based method, DACOOP, we use the same RL backbone, MAPPO, with the same hyperparameters as {\name} to tune the candidate APF parameters.

\begin{figure*}[t]
\begin{minipage}{\textwidth}
\centering
\subfigure[Capture rate.\label{fig:radius_rate}]{\includegraphics[width=0.24\textwidth]{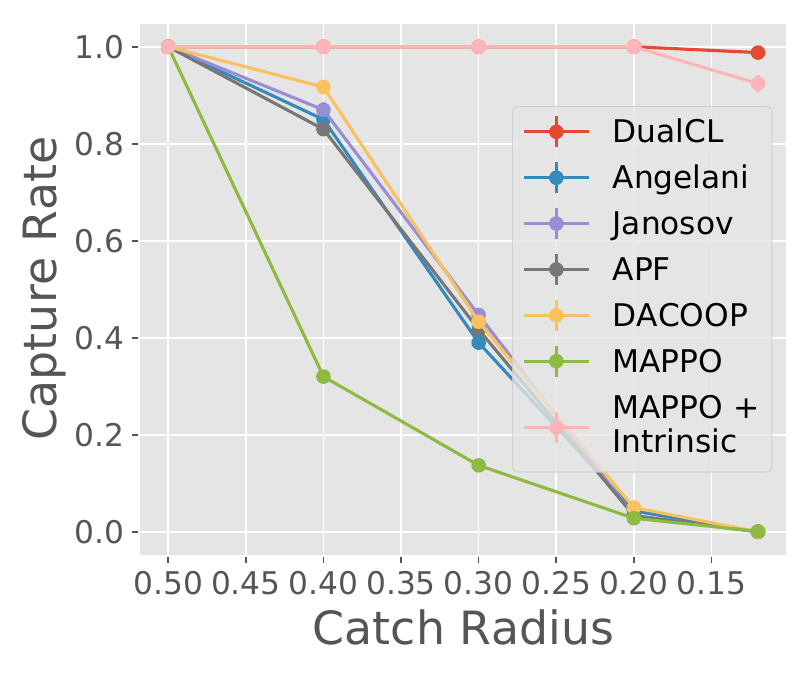}}
\subfigure[Capture timestep.\label{fig:radius_timestep}]{\includegraphics[width=0.24\textwidth]{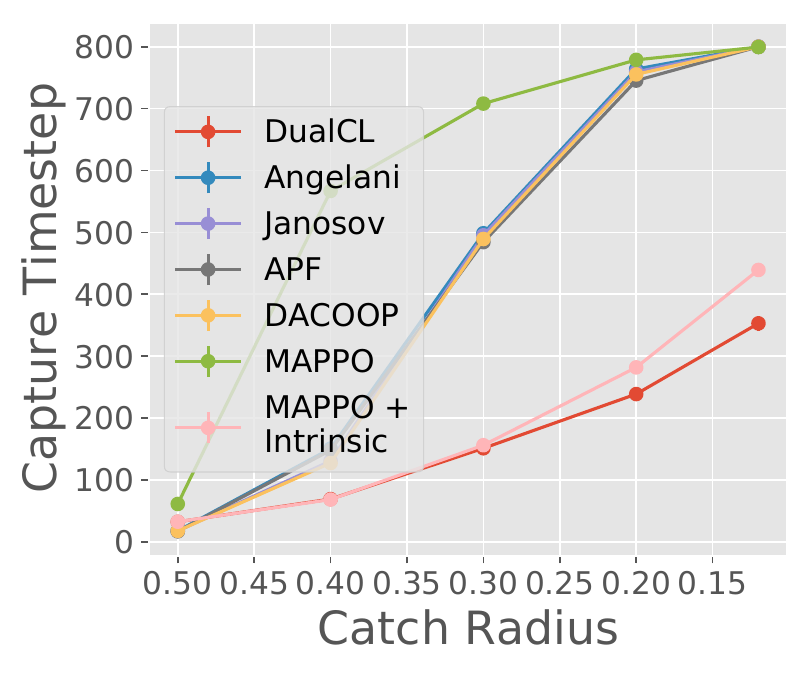}}
\subfigure[Capture rate.\label{fig:speed_rate}]{\includegraphics[width=0.24\textwidth]{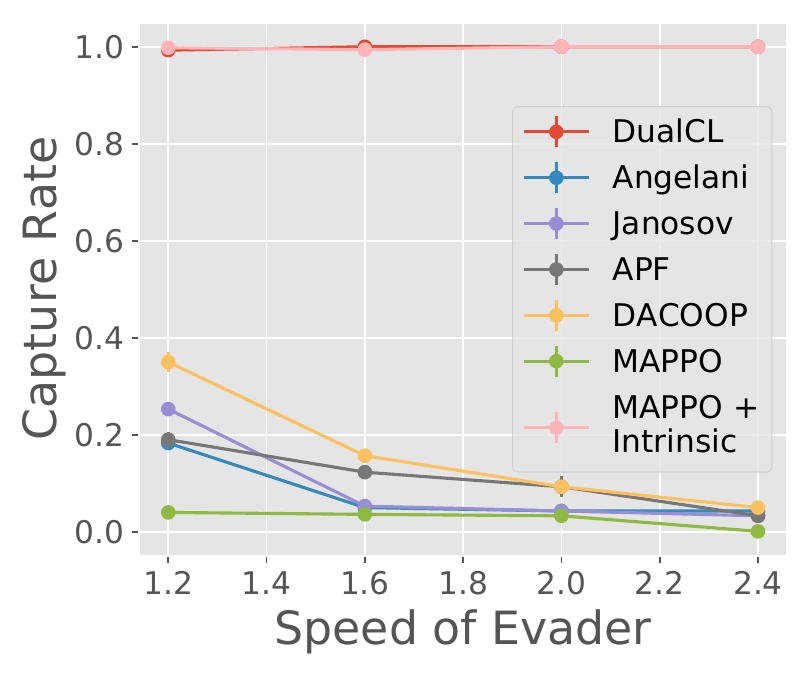}}
\subfigure[Capture timestep.\label{fig:speed_timestep}]{\includegraphics[width=0.24\textwidth]{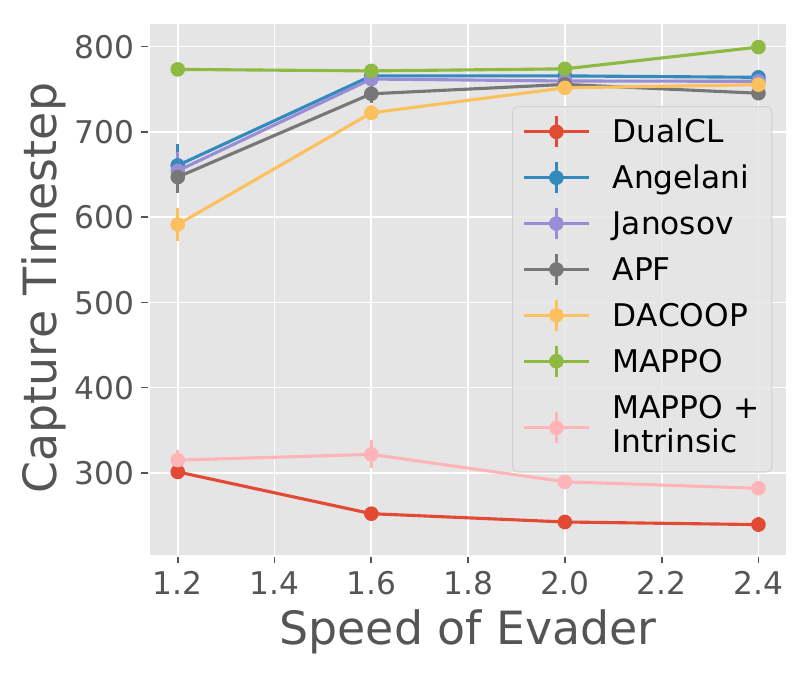}}
\caption{Results of {\name} and all baselines in \emph{Empty}. {\name} achieves $100\%$ capture rate and the shortest capture timestep with all combinations of intrinsic parameters.}\label{fig:empty}
\vspace{-2mm}
\end{minipage}
\end{figure*}

\begin{figure*}[t]
\begin{minipage}{\textwidth}
\centering
\subfigure[Initialization.]{\includegraphics[width=0.13\textwidth]{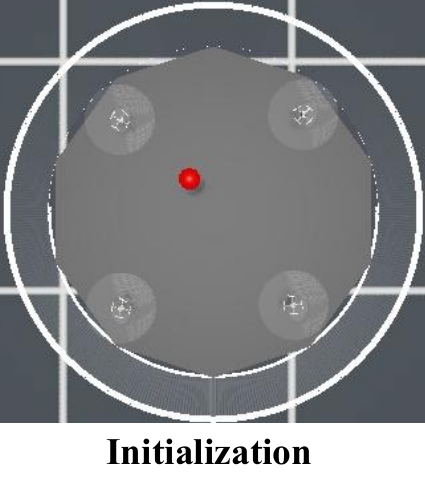}}
\subfigure[APF.\label{fig:heuristic_sim}]
{\includegraphics[width=0.39\textwidth]{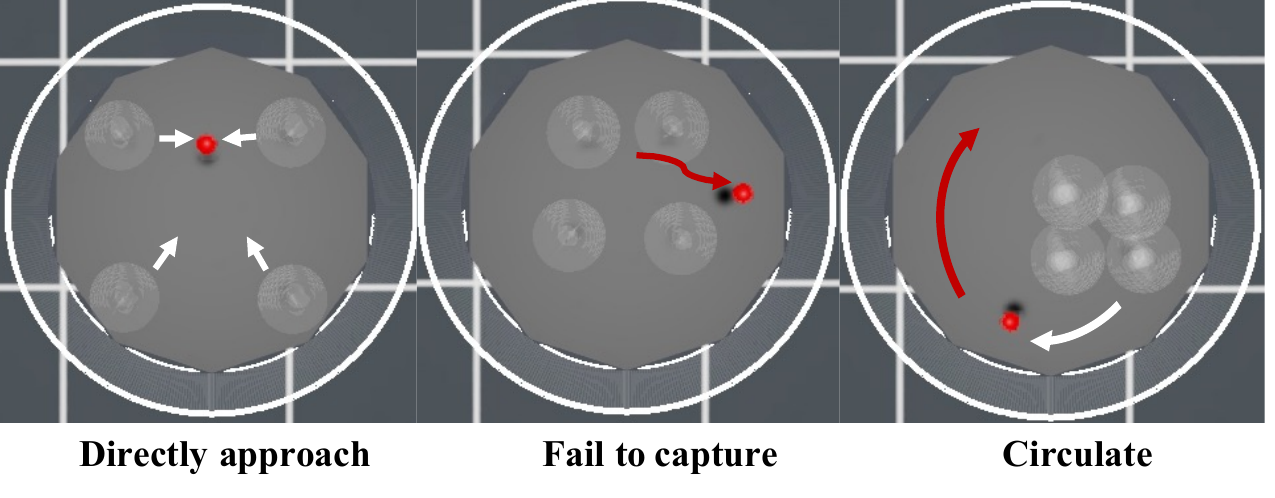}}
\subfigure[{\name}.\label{fig:cl_sim}]{\includegraphics[width=0.39\textwidth]{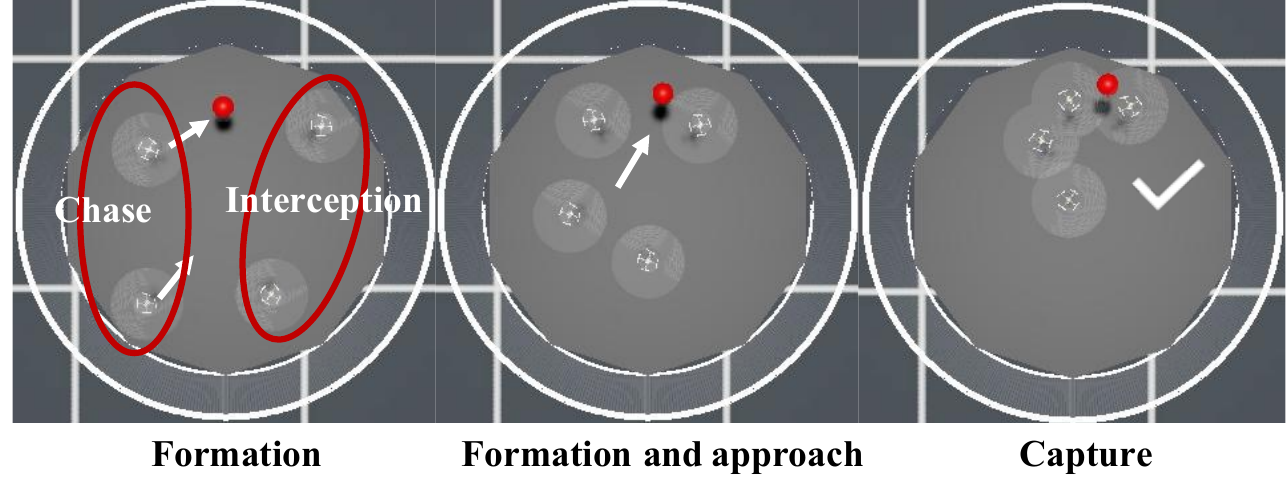}}
\caption{Visualization of the capture strategies generated by {\name} and APF.}
\vspace{-5mm}
\end{minipage}
\end{figure*}

\subsection{Main Results}
As shown in Fig.~\ref{fig:empty}, we compare all algorithms in the \emph{Empty} and illustrate how capture rate and capture timestep change with respect to the intrinsic parameters. Compared to the baselines, our method achieves close to $100\%$ capture rate and the shortest capture timestep with all intrinsic parameters. 
As shown in Fig.~\ref{fig:radius_rate} and~\ref{fig:radius_timestep}, in the \emph{Empty} with the evader's speed $v_e = 2.4$, the capture rate of all heuristic algorithms decreases, and the capture timestep rises rapidly as the catch radius decreases. Heuristic algorithms use a simplified fixed-velocity model to control drones, limiting the expression of complex capture strategies. When the capture radius becomes smaller, the drones need to get closer to the evader, and it is more difficult to form an encircling circle that completely blocks escape routes by hard-coding. 
In Sec.~\ref{sec:visualization}, we also visually analyze why the capture strategies of heuristic algorithms lack cooperation. DACOOP is slightly better than APF because it just adaptively adjusts the hyperparameters of APF depending on the environment and does not enhance the cooperation ability. Meanwhile, MAPPO suffers from serious exploration problems and only works well in scenes with a large catch radius when directly applied to the multi-UAV pursuit-evasion problem. We also evaluate the performance of MAPPO combined with the \emph{Intrinsic Parameter Curriculum Proposer} (MAPPO + \emph{Intrinsic}) and find that its performance is slightly worse than {\name}, which indicates that the \emph{Intrinsic Parameter Curriculum Proposer} can help the RL policy to improve with limited data.

As shown in Fig.~\ref{fig:speed_rate} and~\ref{fig:speed_timestep}, in the \emph{Empty} with the catch radius $d_c = 0.2$, we find that heuristic algorithms and DACOOP perform worse as the speed of the evader grows, mainly due to the fact that the evader can escape from drones faster as the speed increases. MAPPO completely fails, indicating that only decreasing the speed of the evader does not enable RL policy to produce progress when the capture radius is small. MAPPO + \emph{Intrinsic} is the strongest competitor of {\name}, illustrating that the \emph{Intrinsic Parameter Curriculum Proposer} helps the policy insensitive to the increase in the evader's speed.

As shown in Tab.~\ref{tab:main}, we compare {\name} with baselines in test scenarios with the catch radius $d_c = 0.12$ and the speed of the evader $v_e = 2.4$. In the within-distribution scenarios, {\name} obtains a capture rate of over $90\%$ and reduces the capture timestep by at least $27.5\%$. {\name} also exhibits the best zero-shot transfer ability in unseen scenarios.
Heuristic algorithms generally lack cooperation in scenarios with fast prey and a small capture radius. Compared to APF, DACOOP can adjust parameters according to the scenario, yet it fails to address the fundamental issue of insufficient cooperation, resulting in similar performance. We do not list the results of MAPPO in Tab.~\ref{tab:main} because directly applying MAPPO to these scenarios yields no feasible strategies and achieves almost $0$ capture rate due to the high complexity of tasks. MAPPO + \emph{Intrinsic} proves to be a strong baseline, while its zero-shot transfer ability is notably weak.

\subsection{Abaltion Studies}
\label{sec:ab}

To illustrate the effectiveness of the \emph{Intrinsic Parameter Curriculum Proposer} and \emph{External Environment Generator}, we compare four different approaches: MAPPO, MAPPO combined with the \emph{Intrinsic Parameter Curriculum Proposer} (MAPPO + \emph{Intrinsic}), MAPPO combined with the \emph{External Environment Generator} (MAPPO + \emph{External}), and our method, {\name}, in the scenarios where obstacles vary from $0$ to $3$. 

\begin{wrapfigure}{r}{0.25\textwidth}
 \centering
    \includegraphics[width=0.25\textwidth]{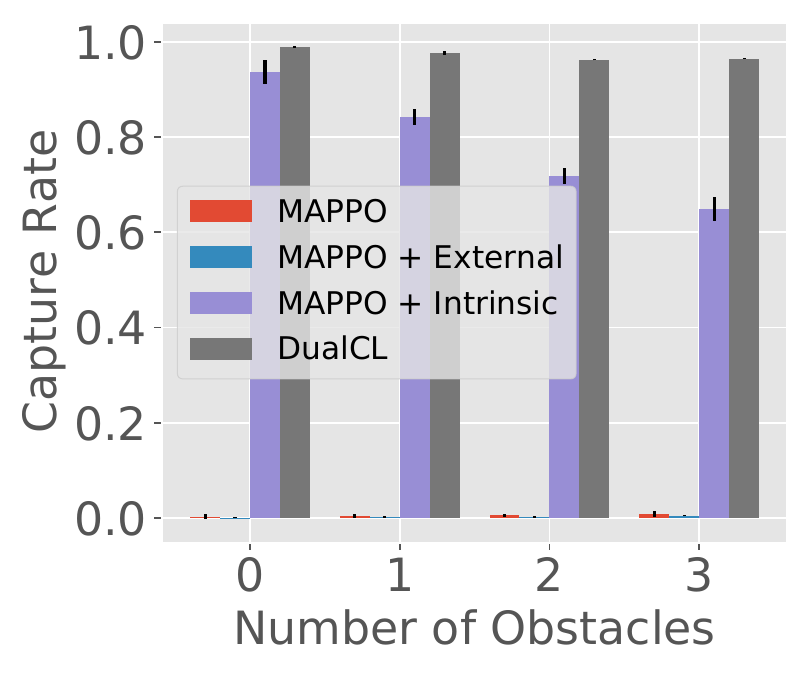}
	\caption{Ablations on two main components.}
 \vspace{-2mm}
\label{fig:ab}
\end{wrapfigure}

As shown in Fig.~\ref{fig:ab}, we compare the capture rate of MAPPO, MAPPO + \emph{Intrinsic}, MAPPO + \emph{External}, and {\name} in the training scenarios with the catch radius $d_c = 0.12$ and speed of the evader $v_e = 2.4$. The results show that MAPPO and MAPPO + \emph{External} fail entirely in the training scenarios with tough conditions, whereas MAPPO + \emph{Intrinsic} provides better performance, which indicates that in complex UAV scenarios, extremely hard intrinsic parameters lead to severe exploration problems, resulting in no improvement for the policy during the early stages of training in RL. 

Through the comparison of MAPPO + \emph{Intrinsic} and {\name}, we find that using the \emph{Intrinsic Parameter Curriculum Proposer} alone causes the policy to converge to the local optima, especially in scenarios with one, two, or three obstacles. The \emph{External Environment Generator} maintains the exploration of unsolved scenes during training scenario generation, increasing the training weights of hard cases and enabling the policy to converge to superior optima.

\subsection{Behavior Analysis and Real-World Deployment}\label{sec:visualization}
We investigate why heuristic algorithms fail when the evader's speed is high, or the capture radius is small. We visualize the capture policy of {\name} and a strong baseline, APF, in the \emph{Empty} with the catch radius $d_c = 0.12$ and the speed of the evader $v_e  = 2.4$. As shown in Fig.~\ref{fig:heuristic_sim}, APF controls the moving direction of drones and does not take into account the acceleration and deceleration, making drones tend to move directly towards the evader. When the evader's speed is high, and the capture radius is small, the capture strategy proposed by APF struggles to adaptively adjust the perimeter to enclose all escape routes of the evader. In contrast, as shown in Fig.~\ref{fig:cl_sim}, we find that the capture strategy in {\name} comprises two key phases: chasing and interception. The drones automatically split into two groups, one chasing the evader from behind, forcing the evader to move. The other group intercepts the evader in advance to prevent the evader from escaping. When the encirclement is complete, the drones keep the formation and approach the evader.

For real-world transfer, we use the \emph{CrazyFlie} 2.1 as quadrotor robots. The maximum speed of quadrotors is limited to $1.0$ m/s. We use a motion capture system to obtain the ground truth of drones’ states. We use a virtual evader and provide the actual states of the evader directly to the RL policy of drones. We infer the RL policy on a local computer and send control commands in the form of collective thrust and desired body rates at $50$ Hz to quadrotors through a radio. Experiments in real-world environments indicate that our method exhibits a similar strategy to the simulation environment, demonstrating that the capture policy generated by {\name} is executable in the real world. The video is available on our website \url{https://sites.google.com/view/dualcl}.  


\section{CONCLUSION}\label{sec:conclusion}
We introduce the dual curriculum learning framework, a novel approach that combines reinforcement learning and curriculum learning to address the multi-UAV pursuit-evasion problem in diverse environments effectively. Our method outperforms existing approaches, consistently achieving over $90\%$ capture rates and the shortest capture timestep. Future work can explore the extension of {\name} to other robotic applications.

\addtolength{\textheight}{-12cm}   





\bibliographystyle{IEEEtran}
\bibliography{references}

\begin{thebibliography}{10}
\providecommand{\url}[1]{#1}
\csname url@rmstyle\endcsname
\providecommand{\newblock}{\relax}
\providecommand{\bibinfo}[2]{#2}
\providecommand\BIBentrySTDinterwordspacing{\spaceskip=0pt\relax}
\providecommand\BIBentryALTinterwordstretchfactor{4}
\providecommand\BIBentryALTinterwordspacing{\spaceskip=\fontdimen2\font plus
\BIBentryALTinterwordstretchfactor\fontdimen3\font minus \fontdimen4\font\relax}
\providecommand\BIBforeignlanguage[2]{{%
\expandafter\ifx\csname l@#1\endcsname\relax
\typeout{** WARNING: IEEEtran.bst: No hyphenation pattern has been}%
\typeout{** loaded for the language `#1'. Using the pattern for}%
\typeout{** the default language instead.}%
\else
\language=\csname l@#1\endcsname
\fi
#2}}

\bibitem{turetsky2003missile}
V.~Turetsky and J.~Shinar, ``Missile guidance laws based on pursuit--evasion game formulations,'' \emph{Automatica}, vol.~39, no.~4, pp. 607--618, 2003.

\bibitem{eklund2011switched}
J.~M. Eklund, J.~Sprinkle, and S.~S. Sastry, ``Switched and symmetric pursuit/evasion games using online model predictive control with application to autonomous aircraft,'' \emph{IEEE Transactions on Control Systems Technology}, vol.~20, no.~3, pp. 604--620, 2011.

\bibitem{oyler2016pursuit}
D.~W. Oyler, P.~T. Kabamba, and A.~R. Girard, ``Pursuit--evasion games in the presence of obstacles,'' \emph{Automatica}, vol.~65, pp. 1--11, 2016.

\bibitem{janosov2017group}
M.~Janosov, C.~Vir{\'a}gh, G.~V{\'a}s{\'a}rhelyi, and T.~Vicsek, ``Group chasing tactics: how to catch a faster prey,'' \emph{New Journal of Physics}, vol.~19, no.~5, p. 053003, 2017.

\bibitem{angelani2012collective}
L.~Angelani, ``Collective predation and escape strategies,'' \emph{Physical review letters}, vol. 109, no.~11, p. 118104, 2012.

\bibitem{fang2020cooperative}
X.~Fang, C.~Wang, L.~Xie, and J.~Chen, ``Cooperative pursuit with multi-pursuer and one faster free-moving evader,'' \emph{IEEE transactions on cybernetics}, vol.~52, no.~3, pp. 1405--1414, 2020.

\bibitem{gupta2017cooperative}
J.~K. Gupta, M.~Egorov, and M.~Kochenderfer, ``Cooperative multi-agent control using deep reinforcement learning,'' in \emph{Autonomous Agents and Multiagent Systems: AAMAS 2017 Workshops, Best Papers, S{\~a}o Paulo, Brazil, May 8-12, 2017, Revised Selected Papers 16}.\hskip 1em plus 0.5em minus 0.4em\relax Springer, 2017, pp. 66--83.

\bibitem{wang2019largescale}
W.~Wang, T.~Yang, Y.~Liu, J.~Hao, X.~Hao, Y.~Hu, Y.~Chen, C.~Fan, and Y.~Gao, ``From few to more: Large-scale dynamic multiagent curriculum learning,'' 2019.

\bibitem{desouky2011q}
S.~F. Desouky and H.~M. Schwartz, ``Q ($\lambda$)-learning adaptive fuzzy logic controllers for pursuit--evasion differential games,'' \emph{International Journal of Adaptive Control and Signal Processing}, vol.~25, no.~10, pp. 910--927, 2011.

\bibitem{awheda2015residual}
M.~D. Awheda and H.~M. Schwartz, ``The residual gradient facl algorithm for differential games,'' in \emph{2015 IEEE 28th Canadian Conference on Electrical and Computer Engineering (CCECE)}.\hskip 1em plus 0.5em minus 0.4em\relax IEEE, 2015, pp. 1006--1011.

\bibitem{wang2020cooperative}
Y.~Wang, L.~Dong, and C.~Sun, ``Cooperative control for multi-player pursuit-evasion games with reinforcement learning,'' \emph{Neurocomputing}, vol. 412, pp. 101--114, 2020.

\bibitem{zhang2022multi}
Z.~Zhang, X.~Wang, Q.~Zhang, and T.~Hu, ``Multi-robot cooperative pursuit via potential field-enhanced reinforcement learning,'' in \emph{2022 International Conference on Robotics and Automation (ICRA)}.\hskip 1em plus 0.5em minus 0.4em\relax IEEE, 2022, pp. 8808--8814.

\bibitem{xu2023omnidrones}
B.~Xu, F.~Gao, C.~Yu, R.~Zhang, Y.~Wu, and Y.~Wang, ``Omnidrones: An efficient and flexible platform for reinforcement learning in drone control,'' \emph{arXiv preprint arXiv:2309.12825}, 2023.

\bibitem{muro2011wolf}
C.~Muro, R.~Escobedo, L.~Spector, and R.~Coppinger, ``Wolf-pack (canis lupus) hunting strategies emerge from simple rules in computational simulations,'' \emph{Behavioural processes}, vol.~88, no.~3, pp. 192--197, 2011.

\bibitem{xu2020multi}
L.~Xu, B.~Hu, Z.~Guan, X.~Cheng, T.~Li, and J.~Xiao, ``Multi-agent deep reinforcement learning for pursuit-evasion game scalability,'' in \emph{Proceedings of 2019 Chinese Intelligent Systems Conference: Volume I 15th}.\hskip 1em plus 0.5em minus 0.4em\relax Springer, 2020, pp. 658--669.

\bibitem{huttenrauch2019deep}
M.~H{\"u}ttenrauch, S.~Adrian, G.~Neumann, \emph{et~al.}, ``Deep reinforcement learning for swarm systems,'' \emph{Journal of Machine Learning Research}, vol.~20, no.~54, pp. 1--31, 2019.

\bibitem{lowe2017multi}
R.~Lowe, Y.~I. Wu, A.~Tamar, J.~Harb, O.~Pieter~Abbeel, and I.~Mordatch, ``Multi-agent actor-critic for mixed cooperative-competitive environments,'' \emph{Advances in neural information processing systems}, vol.~30, 2017.

\bibitem{zhang2022game}
R.~Zhang, Q.~Zong, X.~Zhang, L.~Dou, and B.~Tian, ``Game of drones: Multi-uav pursuit-evasion game with online motion planning by deep reinforcement learning,'' \emph{IEEE Transactions on Neural Networks and Learning Systems}, 2022.

\bibitem{bengio2009curriculum}
Y.~Bengio, J.~Louradour, R.~Collobert, and J.~Weston, ``Curriculum learning,'' in \emph{Proceedings of the 26th annual international conference on machine learning}, 2009, pp. 41--48.

\bibitem{portelas2020automatic}
R.~Portelas, C.~Colas, L.~Weng, K.~Hofmann, and P.-Y. Oudeyer, ``Automatic curriculum learning for deep rl: A short survey,'' \emph{arXiv preprint arXiv:2003.04664}, 2020.

\bibitem{florensa2017reverse}
C.~Florensa, D.~Held, M.~Wulfmeier, M.~Zhang, and P.~Abbeel, ``Reverse curriculum generation for reinforcement learning,'' in \emph{Conference on robot learning}.\hskip 1em plus 0.5em minus 0.4em\relax PMLR, 2017, pp. 482--495.

\bibitem{florensa2018automatic}
C.~Florensa, D.~Held, X.~Geng, and P.~Abbeel, ``Automatic goal generation for reinforcement learning agents,'' in \emph{International conference on machine learning}.\hskip 1em plus 0.5em minus 0.4em\relax PMLR, 2018, pp. 1515--1528.

\bibitem{ivanovic2019barc}
B.~Ivanovic, J.~Harrison, A.~Sharma, M.~Chen, and M.~Pavone, ``Barc: Backward reachability curriculum for robotic reinforcement learning,'' in \emph{2019 International Conference on Robotics and Automation (ICRA)}.\hskip 1em plus 0.5em minus 0.4em\relax IEEE, 2019, pp. 15--21.

\bibitem{akkaya2019solving}
I.~Akkaya, M.~Andrychowicz, M.~Chociej, M.~Litwin, B.~McGrew, A.~Petron, A.~Paino, M.~Plappert, G.~Powell, R.~Ribas, \emph{et~al.}, ``Solving rubik's cube with a robot hand,'' \emph{arXiv preprint arXiv:1910.07113}, 2019.

\bibitem{mehta2020active}
B.~Mehta, M.~Diaz, F.~Golemo, C.~J. Pal, and L.~Paull, ``Active domain randomization,'' in \emph{Conference on Robot Learning}.\hskip 1em plus 0.5em minus 0.4em\relax PMLR, 2020, pp. 1162--1176.

\bibitem{chen2021variational}
J.~Chen, Y.~Zhang, Y.~Xu, H.~Ma, H.~Yang, J.~Song, Y.~Wang, and Y.~Wu, ``Variational automatic curriculum learning for sparse-reward cooperative multi-agent problems,'' \emph{Advances in Neural Information Processing Systems}, vol.~34, pp. 9681--9693, 2021.

\bibitem{long2020evolutionary}
Q.~Long, Z.~Zhou, A.~Gupta, F.~Fang, Y.~Wu, and X.~Wang, ``Evolutionary population curriculum for scaling multi-agent reinforcement learning,'' in \emph{International Conference on Learning Representations}, 2020.

\bibitem{de2021decentralized}
C.~De~Souza, R.~Newbury, A.~Cosgun, P.~Castillo, B.~Vidolov, and D.~Kuli{\'c}, ``Decentralized multi-agent pursuit using deep reinforcement learning,'' \emph{IEEE Robotics and Automation Letters}, vol.~6, no.~3, pp. 4552--4559, 2021.

\bibitem{yu2022mappo}
C.~Yu, A.~Velu, E.~Vinitsky, Y.~Wang, A.~Bayen, and Y.~Wu, ``The surprising effectiveness of ppo in cooperative, multi-agent games,'' \emph{arXiv preprint arXiv:2103.01955}, 2021.

\bibitem{tobin2017domain}
J.~Tobin, R.~Fong, A.~Ray, J.~Schneider, W.~Zaremba, and P.~Abbeel, ``Domain randomization for transferring deep neural networks from simulation to the real world,'' in \emph{2017 IEEE/RSJ international conference on intelligent robots and systems (IROS)}.\hskip 1em plus 0.5em minus 0.4em\relax IEEE, 2017, pp. 23--30.

\bibitem{koren1991potential}
Y.~Koren, J.~Borenstein, \emph{et~al.}, ``Potential field methods and their inherent limitations for mobile robot navigation.'' in \emph{Icra}, vol.~2, no. 1991, 1991, pp. 1398--1404.

\end{thebibliography}

\end{document}